\documentclass[10pt,twocolumn,letterpaper]{article}
\usepackage[pagenumbers]{cvpr} 
\makeatletter
\@namedef{ver@everyshi.sty}{}
\makeatother
\usepackage{pgf}
\usepackage{graphicx}

\usepackage{times}
\usepackage{epsfig}
\usepackage{multirow}
\usepackage{mathtools}
\usepackage{makecell}
\usepackage{xcolor}         
\newcommand{\cmark}{\ding{51}}%
\usepackage{enumitem}
\usepackage[linesnumbered,ruled,vlined]{algorithm2e}
\DontPrintSemicolon

\SetKwComment{Comment}{\color{green!50!black}\#}{}

\SetKwProg{Function}{def }{}{}

\usepackage{pifont}
\newcommand{\myparagraph}[1]{\vspace{1pt} \noindent \textbf{#1} }
\usepackage{amsmath}
\usepackage{amssymb}
\usepackage{booktabs}

\usepackage[pagebackref,breaklinks,colorlinks]{hyperref}
\usepackage[capitalize]{cleveref}
\crefname{section}{Sec.}{Secs.}
\Crefname{section}{Section}{Sections}
\Crefname{table}{Table}{Tables}
\crefname{table}{Tab.}{Tabs.}

\begin{document}
\title{Learning Local-Global Contextual Adaptation for Multi-Person Pose Estimation}

\author{Nan Xue~\textsuperscript{1} \and Tianfu Wu~\textsuperscript{2} \and Gui-Song Xia~\textsuperscript{1} \and Liangpei Zhang~\textsuperscript{3} \and \\
\textsuperscript{1} School of Computer Science, Wuhan University \\
\textsuperscript{2} Department of ECE, NC State University \\
\textsuperscript{3} LIESMARS, Wuhan University
}

\maketitle

\begin{abstract}\vspace{-10pt}
    This paper studies the problem of multi-person pose estimation in a bottom-up fashion.
    With a new and strong observation that the localization issue of the center-offset formulation can be remedied in a local-window search scheme in an ideal situation, we propose a multi-person pose estimation approach, dubbed as LOGO-CAP, by learning the LOcal-GlObal Contextual Adaptation for human Pose. Specifically, our approach learns the keypoint attraction maps (KAMs) from the local keypoints expansion maps (KEMs) in small local windows in the first step, which are subsequently treated as dynamic convolutional kernels on the keypoints-focused global heatmaps for contextual adaptation, achieving accurate multi-person pose estimation. 
    Our method is end-to-end trainable with near real-time inference speed in a single forward pass, obtaining state-of-the-art performance on the COCO keypoint benchmark for bottom-up human pose estimation. With the COCO trained model, our method also outperforms prior arts by a large margin on the challenging OCHuman dataset. 
\end{abstract}

\vspace{-15pt}
\section{Introduction}\label{sec:introduction}
\vspace{-2mm}
2D human pose estimation is a classical computer vision problem that aims to parsing articulated structures of human parts from natural images. With rich and longstanding studies, we have witnessed great successes on single-person pose estimation~\cite{SHG,HRNet} by convolutional neural networks. Therefore, it is of great interest in pushing the pose estimation from the single-person to multi-person configuration.

\begin{figure}[!t]
    \centering
    \includegraphics[width=0.85\linewidth]{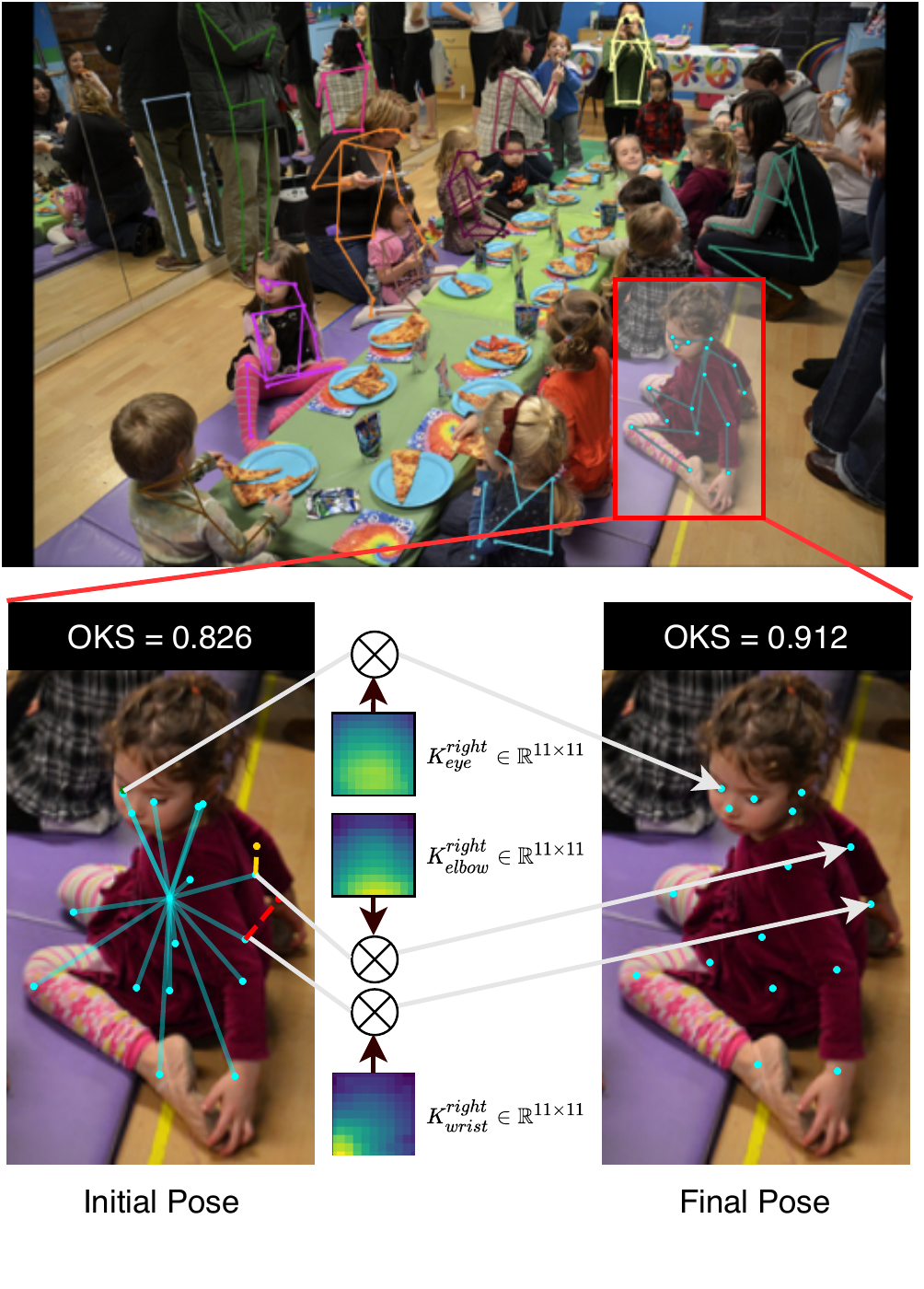}
    \vspace{-5mm}
    \caption{An illustrative example of multi-person pose estimation by the proposed LOGO-CAP with the HRNet-W32 backbone. For each initial pose obtained by the center-offset regression, LOGO-CAP learns $11\times 11$ local filters for each joint, and then refines the initial keypoint by convolution with the learned kernels: \textit{The local filters are learned to \textbf{refocusing} those initially-less-accurate pose keypoints towards better placement.} 
    In more detailed, we show an example of the initial center-offset pose that only obtains the OKS of 0.826 due to the misplacement between the keypoints of {\tt the right elbow} and {\tt the right wrist}. We mark the residual vectors between the predictions and the groundtruth with yellow and red dash lines respectively. The LOGO-CAP improves the OKS by 10.4\%. Please see text for details.
    }
    \vspace{-5mm}
    \label{fig:teaser}
\end{figure}

The problem of multi-person pose estimation from images has been extensively studied in top-down paradigms~\cite{HRNet,SimpleBaseLine,SHG} that formulate the problem as single-person pose estimation with an off-the-shelf person detector with an impressively high AP (e.g, 56\% on the COCO-2017 validation dataset~\cite{SimpleBaseLine}). 
Although the top-down paradigm has been dramatically pushed into a high-performing stages, it remains several problems in both aspects of efficiency and accuracy due to the dependency of detecting person bounding boxes. Motivated by this, we are interested in studying the problem of multi-person pose estimation without incurring extra priors of bounding boxes, which is conventionally termed as \emph{bottom-up pose estimation}.

\begin{figure}[t!]
    \resizebox{0.95\linewidth}{!}{
    \input{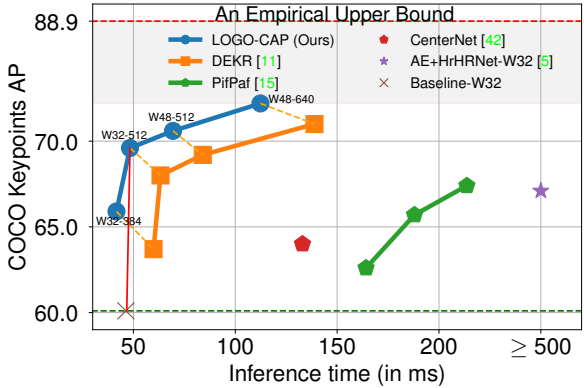}
    }
    \vspace{-3mm}
    \caption{Our LOGO-CAP is motivated by a strong empirical observation that a vanilla center-offset baseline with an AP of 60.1 (marked by \textcolor{green}{green} dashed line) can be improved by leveraging a searching scheme in $11\times 11$ local windows to an AP of 88.9 (marked by \textcolor{red}{red} dashed line). In the meanwhile, we illustrate the speed-accuracy comparisons between our LOGO-CAP and prior arts on the COCO val-2017 dataset. W$x$-$Y$ (e.g. W32-384) means that a model uses the backbone HRNet-W$x$~\cite{HRNet} and is tested with the image resolution $Y$ in the short side.
}
    \label{fig:speed-accuracy}\vspace{-4mm}
\end{figure} 

Bottom-up pose estimation approaches pay much attention on estimating pose parameters from any given multi-person image instead of using the cropped single-person ones, which poses several challenges on accurately identifying the learned bottom-level keypoints as person parts and grouping them into different individuals by learning part affinity fields~\cite{paf,openpose}, part association fields~\cite{pifpaf}, associative embedding~\cite{asso_embedding}. Those grouping approaches are accurate but sophisticated due to the necessity of ad-hoc grouping/decoding schemes. 
Recently, many researchers attempted to learning  center-offsets~\cite{centernet-zhou,CenterNet-duan,ART-pose,DEKR-pose} for pose estimation as its intrinsic simplicity and efficiency. However, 
most of the center-offset approaches are suffering from the main challenge of localization inaccuracy due to the large structural variations of human pose, thus leading to inferior performance than the grouping ones.

In this paper, we are interested in studying the bottom-up pose estimation using the center-offset formulation for its simplicity and efficiency. We directly address the aforementioned main challenges for center-offset based multi-person pose estimation. Our proposed approach is motivated by a surprisingly strong empirical observation via analyzing what the fundamental issues are with the vanilla center-offset pose estimation network.

\myparagraph{A Surprisingly Strong Observation.} 
Our analyses are based on results in the fully-annotated subset of the COCO val-2017 dataset\footnote{Note that the COCO-val-2017 dataset contain many partially-annotated images (with only ground-truth bounding boxes), we use 2346 testing samples that are fully annotated with keypoint annotations.}. 
The vanilla center-offset regression method utilizes the HRNet-W32~\cite{HRNet} as the feature backbone to directly predict keypoints center heatmap and the offset vectors. As shown in \cref{fig:speed-accuracy}, it obtains $60.1$ average precision (AP), which is not great, but reasonably good. It clearly shows that the pose keypoints center and the offset vectors can be learned reasonably well. We ask: \textit{How exactly bad are the center-offset estimation?} We want to know (i) whether some of keypoints are truly bad being far away from the underlying ground-truth (GT) locations, or (ii) whether most of them are already in the close proximity of the GT ones?  We observe that the latter is true. To quantitatively characterize the close proximity, instead of directly utilizing the learned offset vectors for human pose estimation, we treat them as human pose keypoint initialization and do a local window search to compute the empirical upper-bound of performance. More specifically, based on the initially predicted human poses, by introducing a local window (e.g., $11\times 11$) centered at each detected key point and by computing the single keypoint similarity with the ground-truth keypoint, \textbf{an empirical upper-bound of $88.9$ AP is obtained}, which is significantly higher than the state of the art and shows the potential of improving the vanilla center-offset regression paradigm.

\textbf{A Direct Solution Leveraging the Observation}. The implication of the observation is significant: It reveals that the fundamental challenge of improving the center-offset approach for pose estimation is to resolve the local misplacement.   
To that end, a straightforward way is just to learn a local heatmap (e.g., $11\times 11$) for each human pose keypoint based on the learned center and offset vectors, and then to compute the refined keypoints by taking $\arg \max$ within the local heatmap. Although appealing, this does not work as observed during our development of the LOGO-CAP. The underlying reason is also straightforward: if this can work, the original offset vector regression should work at the first place since no additional information is introduced through learning the local heatmap.

\textit{We hypothesize} that on the one hand, in addition to the local heatmap, the structural relationship between different pose keypoints needs to be taken into account, and on the other hand, the intrinsic uncertainty of the local information in a local heatmap needs to be resolved. The former  is the key challenge of structured output prediction problems. Many message passing algorithms have been developed in the literature. The latter can not be addressed by simply increasing the local window size. It entails learning stronger local-global information interaction and adaptation.

To verify the two hypotheses, the proposed LOGO-CAP lifts the initial keypoints via the center-offset prediction to keypoint expansion maps (KEMs) to counter their lack of localization accuracy in two modules (Section~\ref{sec:LOGO-CAP}). The KEMs extend the star-structured representation of the center-offset formulation to the pictorial structure representation~\cite{fischler1973representation,felzenszwalb2005pictorial}. As shown in \cref{fig:teaser}, in our LOGO-CAP, one module computes local KEMs and learns to account for the structured output prediction nature of the human pose estimation problem, resulting in the keypoint attraction maps (KAMs), i.e., the local filters in \cref{fig:teaser}. Another module computes global KEMs and learns to refine the global KEMs by integrating the KAMs. 

Our LOGO-CAP is a fully end-to-end bottom-up human pose estimation method with near real-time inference speed. It obtains $70.0$ AP in the fully-annotated subset of the COCO val-2017 dataset, which is an absolute increase of $9.9$ AP compared to the vanilla center-offset method, making a significant step forward. 
Fig.~\ref{fig:speed-accuracy} shows the advantage of the proposed LOGO-CAP in terms of overall speed-accuracy comparisons between our LOGO-CAP and prior arts. Meanwhile, we should notice that there is still a significant gap towards the empirical upper bound in \cref{fig:speed-accuracy}, which encourages more work to be investigated.    
\section{Related Works and Our Contributions}\vspace{-2mm}
There is a vast body of literature for human pose estimation. Many elegant representation schema have been developed for modeling articulated human pose in the traditional approaches such as the well-known pictorial structure model~\cite{fischler1973representation,felzenszwalb2005pictorial} and its many variants~\cite{ramanan2005strike, andriluka2009pictorial,pishchulin2013poselet,yang2012articulated,rothrock2013integrating}.  With the resurgence of DNNs and the end-to-end learning, the performance of single-person pose estimation has been largely improved. 
We briefly review the recent deep learning based approaches for multi-person pose estimation. 

\myparagraph{Top-Down Pose Estimation.} It essentially exploits single-person pose estimation  approaches on cropped single human image patches ~\cite{SHG,HRNet,SimpleBaseLine,Topdown-CPN,Topdown-RMPE,Topdown-G-RMI,Topdown-PRTR}, where human detection is often done by an off-the-shelf object detector (\eg, Faster-RCNN~\cite{Faster-RCNN}). Although excellent performance has been achieved in such a pipeline, they are suffering from efficiency issue due to the dependency of object detectors. Accordingly, there are work focusing on developing efficient backbone networks (\eg, Lite-HRNet~\cite{Lite-HRNet, MobileNetV2, ShuffleNetV2}) to reduce the inference latency on single-person pose estimation, often at the expense of degenerating the accuracy of pose estimation by large margins. Mask-RCNN~\cite{Mask_RCNN} addresses this problem by learning heatmaps from features extracted by ROIAlign in object proposals, whose performance on pose estimation lags behind. 
To clarify the use of terminology, top-down approaches referred in this paper are those using the precomputed bounding boxes as done in the SimpleBaseline method~\cite{SimpleBaseLine}. Compared with the top-down approaches, our LOGO-CAP directly addressed the problem of multi-person pose estimation in a bottom-up way without incurring the regional image/feature context over the person boxes. Our proposed LOGO-CAP obtains competitive performance in terms of accuracy while achieving nearly real-time inference speed.

\myparagraph{Limb-based Grouping Approaches.} Motivated by the naturalness of modeling limbs based on keypoints, the problem of multi-person pose estimation has been extensively studied by grouping persons from the learned limb associations. Given a predefined limb configuration (\eg, the COCO person skeleton template consisting of 19 limbs based on 17 keypoints), the grouping can be addressed by Part affinity field (PAF)~\cite{paf,openpose}, Associative Embedding (AE)~\cite{asso_embedding}, mid-range offset fields in PersonLab~\cite{personlab} and the fields of Part Intensity and Association~\cite{pifpaf}. Typically, sophisticated designs are entailed to achieve good performance. For example, a bipartite graph matching is used in OpenPose~\cite{openpose}.
In addition to be computationally expensive, another drawback of these methods is not fully end-to-end trainable. More recently, the differentiability issue was studied by the Hierarchical Graph Clustering (HGG) method~\cite{HGG}, which utilizes graph convolution networks to repeatedly delineate pose parameters of multiple persons from a keypoint graph. HGG improves the performance compared to its baseline, the Associative Embedding method~\cite{asso_embedding} at the expense of significantly increased computational cost. In contrast to those approaches, our proposed LOGO-CAP is fully end-to-end trainable and achieves near real-time inference speed. 

\begin{figure*}[!t]
    \centering
    \includegraphics[width=0.9\linewidth]{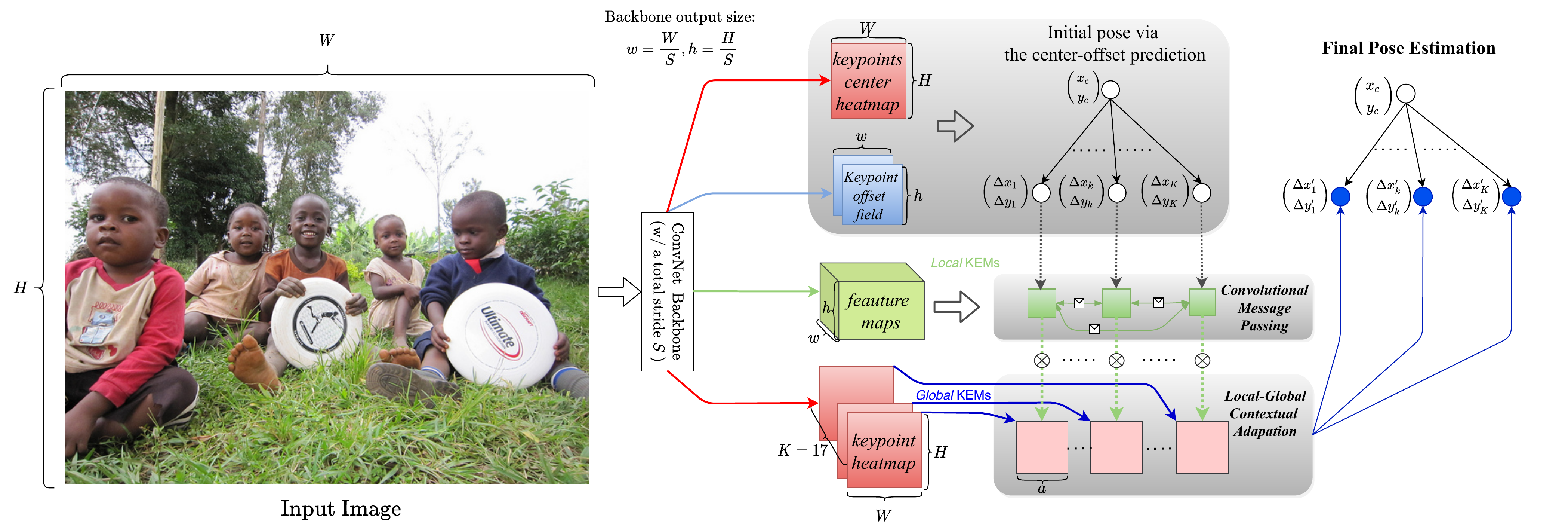}
    \vspace{-3mm}
    \caption{Schematic illustration of the proposed LOGO-CAP for bottom-up human pose estimation. See text for detail.}
    \label{fig:LOGO-CAP}\vspace{-3mm}
\end{figure*} 

\myparagraph{Direct Regression based Approaches.} This formulation has attracted much attention due to their conceptual simplicity~\cite{CenterNet-duan,direct-pose,ART-pose,DEKR-pose,mixture-dense-pose}, inspired by  the recent remarkable success of direct bounding box regression in object detection such as the FCOS method~\cite{FCOS} and CenterNets~\cite{centernet-zhou,CenterNet-duan}. As aforementioned, one main challenge is the difficulty of accurately regressing the offset vectors, especially for the long-range keypoints with respect to the center. Sophisticated post-processing schema are often entailed to improve the performance. For example, a method of matching the directly regressed poses to the nearest keypoints that are extracted from the global keypiont heatmaps is used in~\cite{centernet-zhou}. 
Although being simple, the performance of this line of work is usually inferior to the limb-based approaches. The mixture regression network~\cite{mixture-dense-pose} alleviated the issue of regression quality to some extent, but still remained an indispensable performance gap comparing with the grouping-based approaches.  
Most recently, Geng \emph{et al.} presented the first competitive direct method, DEKR~\cite{DEKR-pose} with a novel pose-specific neural architecture for disentangled keypoint regression. To improve the performance, the DEKR method utilizes a lightweight rescoring network to recalibrate the pose scores that are computed based on the keypoint heatmaps
, and thus is not fully end-to-end. The proposed LOGO-CAP retains the simplicity of the vanilla center-offset formulation and enjoys fully end-to-end training and fast inference speed.

\myparagraph{Our Contributions.} The proposed LOGO-CAP makes three main contributions to the field of bottom-up human pose estimation: (i) We address the drawback of the vanilla center-offset formulation while retaining its efficiency. It proposes the key idea of lifting a keypoint to a keypoint expansion map to counter the lack of localization accuracy. 
(ii) We present a novel local-global contextual adaptation formulation that accounts for the nature of structured output prediction in human pose estimation and harnesses local-global structural information integration.
(iii) Our proposed LOGO-CAP obtains state-of-the-art performance on the COCO val-2017 and test-2017 datasets. It also shows the strong generalization ability with state-of-the-art performance on the OCHuman dataset.


\section{The Proposed LOGO-CAP}\label{sec:LOGO-CAP}\vspace{-2mm}
\cref{fig:LOGO-CAP} illustrates the proposed LOGO-CAP, which consists of three main components: a feature backbone, the initial center-offset human pose estimation, and the proposed local-global contextual adaptation component for the final human pose estimation. 


\subsection{Learning Local and Global Context}\vspace{-1mm}
\myparagraph{Backbone Network and Pose Initialization.}
Given an input image $I$, the output of the feature backbone is a $C$-dim feature map, denoted by $\mathbf{F}\in R^{C\times h \times w}$, where $C$ is the feature dimension of the last convolutional layer in the feature backbone, and the spatial size $h\times w$ depends on the total stride in the feature backbone. The feature map $F$ is used as the shared features by two different head branches for the prediction of center heatmaps $\mathbf{C}$ and offsets fields $\mathbf{O}$. Then, the initial pose parameters are extracted using the top-$N$ local maximal locations and the corresponding offset vectors, denoted by $\mathbf{p}_i \in\mathbb{R}^{17\times 2}$ for the $i$-th person.

\myparagraph{Local Context.} We first introduce the local keypoint expansion maps (KEMs) for the initial pose parameters. Specifically, for the $j$-type keypoint (\eg, nose of a person), we follow the \textbf{strong observation} in \cref{sec:introduction} to compute the local KEMs in  $11\times 11$ windows as $\mathcal{M}_{N\times 17\times 11\times 11\times 2}$, as shown in \cref{alg:kpt-expansion}. 
\begin{algorithm}[!t]
\small{
  \caption{\small Computing KEMs in a PyTorch-like style}\label{alg:kpt-expansion}
  
  $\sigma$= coco\_sigmas 
  \Comment{$17\times 1$, keypoint sigmas provided in the COCO dataset.}

  \Function{KptsExpansionCoco($\mathcal{P}$,ks=11)}
  {
  \Comment{Initial poses $\mathcal{P}$: Nx17x2}
  r = ks // 2
  
  dy, dx = meshgrid(arange(-r,r),arange(-r,r))
  
  \Comment{dx, dy: ks$\times$ks}
  
  dy = dy.reshape(1, 1, ks, ks)

  dx = dx.reshape(1, 1, ks, ks)
  
  scale = $\sigma$.reshape(1, 17, 1, 1)/$\sigma$.min()
  \Comment{keypoint type specific expansion rate}
  
  dy = dy $*$ scale \Comment{ 1x17x11x11}
  
  dx = dx $*$ scale \Comment{ 1x17x11x11}
  
  dxy = stack((dx,dy),dim=-1) \Comment{ 1x17x11x11x2}
  
  $\mathcal{M}$ = $\mathcal{P}$.reshape(N, 17, 1, 1, 2) + dxy \Comment{  Nx17x11x11x2}
  
  \Return $\mathcal{M}$
  }}
  
\end{algorithm}

Then, we encode the geometric mesh of KEMs $\mathcal{M} \in \mathbb{R}^{N\times 17\times 11\times 11\times 2}$ in a $d$-dim latent space (\eg, $d=64$ in our experiments), computed based on the feature backbone output. A pose instance is represented by concatenating all the 17 keypoints. All initial keypoint-based poses are geometrically ``expanded / lifted" and feature-activated, resulting the initial local context, $\mathcal{K} \in \mathbb{R}^{N\times (17\times d)\times 11\times 11}$.

\myparagraph{Local Context Convolutional Message Passing.} To facilitate the structural information flow between different latent codes of the keypoints of a pose instance, we propose a simple convolutional message passing (CMP) module with three layers of Conv+Norm+ReLU operations with the Attentive Norm~\cite{AttnNorm} used in the second layer. 
The transformed latent code $\mathcal{K}'$ is decoded by a $1\times 1$ Conv. layer as the local keypoints attraction maps (KAMs), $\mathbf{K}\in\mathbb{R}^{N\times 17\times 11\times 11}$ to measure the uncertainty of the initial poses. 

\myparagraph{Local-Global Contextual Adaptation.} Through the CMP, we obtain the dynamic (a.k.a., data-driven) kernels for the 17 keypoints in a pose instance-sensitive way, which are used to refine the global heatmaps $\mathcal{H}$ for the 17 keypoints. In detail, we first compute another geometric mesh with window $a\times a$ (e.g., $a=97$) for each keypoint of the $N$ pose instances, and the entire mesh is denoted by $\mathcal{M}_{G} \in \mathbb{R}^{N\times 17\times a\times a \times 2}$. The mesh can be interpreted as the global KEM. It is then instantiated with appearance features extracted from the global heatmaps, and we have,
\begin{align}
     \mathcal{H}^{\uparrow}_{(1:17)} \xRightarrow[\text{bi-linear}]{\mathcal{M}_G} \mathbb{H}
     \xRightarrow[\text{reweighing}]{\mathcal{G}_{a\times a}(0, \sigma)} \bar{\mathbb{H}}\in\mathbb{R}^{N\times 17 \times a \times a} \label{eq:heatmap},
\end{align}
where to encode the Gaussian prior of keypoint heatmaps, the resulting pose-guided heatmaps $\mathbb{H}$ is reweighed by a Gaussian kernel $\mathcal{G}_{a\times a}(0, \sigma=\frac{a-1}{2\times 3})$ (e.g., $\sigma=16$ when $a=97$) in an element-wise way. By doing so, it means that the enlarged mesh follows the $3\sigma$ principle.

Then, we apply the learned keypoint $11\times 11$ kernels $K_{n, i}$'s to convolve the reweighed $a\times a$ heatmap $\bar{\mathbb{H}}_{n, i}$ in a pose instance-sensitive and keypoint-specific way, leading to \textbf{LO}cal-\textbf{G}l\textbf{O}bal \textbf{C}ontextual \textbf{A}daptation, 
\begin{equation}
    \bar{\mathbb{H}} \xRightarrow[\text{LOGO-CA}]{K_{N\times 17 \times 11 \times 11}} \Tilde{\mathbb{H}}\in\mathbb{R}^{N\times 17 \times a \times a}, \label{eq:refinedheatmap}
\end{equation}
which represents the refined heatmaps for the $17$ human pose keypoints.

\myparagraph{The Pose Estimation Output.} With the local-global contextually adpated heatmaps $\Tilde{\mathbb{H}}_{N\times 17 \times a \times a}$, we maintain the top-2 locations for each keypoint within the $a\times a$ heatmap, and then utilize a convex average of the top-2 locations as the final predicted offset vectors (i.e. $(\Delta x'_i, \Delta y'_i)$'s in \cref{fig:LOGO-CAP}), and of their confidence scores as the prediction score,  with a predefined weight $\lambda$ for the top-1 location ($0.75$ in our experiments). Together with the predicted keypoints centers $\mathcal{C}_{N\times 3}$, the final prediction score for each keypoint is the product between the convex average confidence score and the center confidence score. We keep the keypoints whose final scores are greater than $0$. We have,
\begin{equation}
    \{\mathcal{C}_{N\times 3}, \Tilde{\mathbb{H}}\} \xRightarrow[\text{Score thresholding}]{\text{Output}} \{ \hat{L}^n_I; n=1, \cdots N'\}, \label{eq:poseOutput}
\end{equation}
where $N'$ is the number of the final predicted pose instances in an image $I$.

\subsection{Loss Functions in Training}\vspace{-1mm}
In the fully end-to-end training, we need to define loss functions for the global heatmap $\mathcal{H}$, the refined local heatmap $\tilde{\mathbb{H}}$, the offset field $\mathcal{O}$,  and the keypoint kernels. 

\myparagraph{The Heatmap Loss.} The widely adopted mean squared error (MSE) loss is used. Denoted by  $\mathcal{H}^{GT}_{18\times h\times w}$ the ground truth heatmaps in which each keypoint (including the center) is modeled by a 2-D Gaussian with dataset-provided mean and variance. Let $\mathbf{p}=(i, \mathbf{x})$ be the index of the domain $D$ of dimensions ${18\times h\times w}$. For the predicted heatmaps $\mathcal{H}_{18\times h\times w}$, the MSE loss is defined by,
\begin{equation}
    \mathcal{L}_{\mathcal{H}} = 1/|D| \cdot \sum_{\mathbf{p}\in D}\|w(\mathbf{x}) (\mathcal{H}(\mathbf{p}) - \hat{\mathcal{H}}(\mathbf{p}))\|_2^2,
\end{equation}
where $w(\mathbf{x})$ represents the weight for the foreground and the background pixels. The foreground mask is provided by the dataset annotation. In our experiment, we set $w(\mathbf{x}) = 1 / 0.1$ for a foreground / background pixel respectively.

In defining the loss function $\mathcal{L}_{\tilde{\mathbb{H}}}$ for the refined local heatmap $\tilde{\mathbb{H}}$ (Eqn.~\ref{eq:refinedheatmap}), the ground-truth heatmap $\tilde{\mathbb{H}}^{GT}$ is generated on-the-fly based on the mesh $\mathcal{M}^G_{N\times 17\times a \times a \time 2}$ (Eqn.~\ref{eq:heatmap}) and the ground-truth keypoints using a Gaussian model with mean being the displacement between the current predicted keypoints and the ground-truth ones,  and variance $\sigma$ (i.e., the standard deviation of the reweighing Gaussion prior model in Eqn.~\ref{eq:heatmap}). 

\myparagraph{The Offset Field Loss.} The widely adopted SmoothL1 loss~\cite{Faster-RCNN} is used. Let $\mathcal{O}^{GT}\in\mathbb{R}^{34\times h\times w}$ be the ground-truth offset field, and $\mathcal{C}^{GT}$ be the non-empty set of ground-truth keypoints centers.   For the predicted offset field $\mathcal{O}$, we have,
\begin{equation}
    \mathcal{L}_{\mathcal{O}}(\mathbf{p}) = \mathcal{A}(\mathbf{p}) \ell_1^{\text{smooth}} \left(\mathcal{O}(\cdot, \mathbf{p}), \mathcal{O}^{GT}(\cdot, \mathbf{p}); \beta\right),
\end{equation}
for each foreground pixel $\mathbf{p} \in \mathcal{C}^{GT}$, and
\begin{equation}
    \mathcal{L}_{\mathcal{O}} = \frac{1}{|\mathcal{C}^{GT}|} \cdot \sum_{\mathbf{p}\in\mathcal{C}^{GT}} \mathcal{L}_{\mathcal{O}}(\mathbf{p})
\end{equation}
where $\mathcal{A}(\mathbf{p})$ is the area of the person centered at the pixel $\mathbf{p}$, and $\beta$ the cutting-off threshold (e.g., $\frac{1}{9}$ in our experiments).

\myparagraph{The OKS Loss for the Keyoint Kernels.} Consider a single predicted pose instance, learning the keypoint kernels, $K_{17\times 11\times 11}$ is the key to facilitate the local-global contextual adaptation. To that end, the figure of merits of the KEMs, $\mathcal{M}_{17\times 11 \times 11 \times 2}$ needs to directly reflect the task loss, i.e., the OKS loss.  With respect to the $N^{GT}$ ground-truth pose instances in an image, we can compute the similarity score per keypoint candidate in the KEMs, and obtain the score tensor $S_{17\times 11 \times 11 \times N^{GT}}$. The score tensor is further clamped with a threshold $0.5$, i.e., $S_{17\times 11 \times 11 \times N^{GT}}=\max (S_{17\times 11 \times 11 \times N^{GT}}, 0.5)$. A mean reduction is applied to the first three dimensions of the clamped score tensor to compute the matching score for each of the $N^{GT}$ pose instance. Then, the best ground-truth pose instance indexed by $n^*$ is selected in terms of the matching score, and its matching score is denoted by $s_{n^*}$. Based on the selected ground-truth pose instance, we compute the per-keypoint similarity score for the predicted pose instance at hand, denoted by $s_k$ ($k\in [1, 17]$). Then, the loss function fo the keypoint kernels are defined by, 
\begin{equation}
    \mathcal{L}_{K} = s_{n^*}\cdot \sum_{k,i,j} s_{k} \cdot |K_{k,i,j} - {S}_{k,i,j,n^*}|^2.
\end{equation}

\myparagraph{The Total Loss} is then defined by $\mathcal{L}= \mathcal{L}_{\mathcal{H}} + \mathcal{L}_{\tilde{\mathbb{H}}} + \lambda \cdot (\mathcal{L}_{\mathcal{O}} + \mathcal{L}_{K})$, where the trade-off parameter $\lambda$ is used to balance the different loss items ($\lambda = 0.01$ in our experiments). 
\section{Experiments}\vspace{-2mm}

\begin{table*}[!t]
    \centering
    \vspace{-3mm}
    \caption{Evaluation results on the \texttt{COCO-val-2017} and \texttt{COCO-testdev-2017} dataset. 
    For HGG~\cite{HGG} and SimplePose~\cite{SimplePose}, the multi-scale inference$^\dagger$ is applied on the testdev-2017 dataset. For DEKR~\cite{DEKR-pose} that uses an rescoring network to get the final predictions, we report both the performance with and without rescoring (which is the fair baseline for our LOGO-CAP). The numbers of SPM~\cite{spm} and HGG~\cite{HGG} are extracted from their papers.}
    \vspace{-3mm}
    \resizebox{0.93\textwidth}{!}{
    \begin{tabular}{c|c|c|c|c|c|c|c|c|c|c|c|c}
    \toprule
        & & & \multicolumn{5}{c|}{COCO-val-2017} & \multicolumn{5}{c}{COCO-testdev-2017} \\
        & Method & Backbone  & AP\,[\%]& AP$^{50}$\,[\%] & AP$^{75}$\,[\%] & AP$^M$\,[\%] & AP$^L$\,[\%] & AP\,[\%] & AP$^{50}$\,[\%] & AP$^{75}$\,[\%] & AP$^M$\,[\%] & AP$^L$\,[\%]\\\midrule
        \multirow{3}{*}{\rotatebox{90}{\footnotesize{Top-Down}}}& HRNet~\cite{HRNet}& \scriptsize{HRNet-W48}  & \textcolor{blue}{76.3} &	\textcolor{blue}{90.8}	& \textcolor{blue}{82.9} & \textcolor{blue}{72.3} & \textcolor{blue}{83.4} & \textcolor{blue}{75.5} & 	\textcolor{blue}{92.5} & \textcolor{blue}{83.3}	 & \textcolor{blue}{71.9}	& \textcolor{blue}{81.5}	\\
        & Lite-HRNet~\cite{Lite-HRNet} & \scriptsize{Lite-HRNet-30} & 70.4	& 88.7 & 77.7 & 76.2 & 92.8 & 69.7 & 90.7 & 77.5 & 66.9 & 75.0 \\
        & Mask-RCNN~\cite{Mask_RCNN} & \scriptsize{ResNet-50-FPN} & 64.2 & 86.6 & 69.7 & 58.7 & 73.0 & 63.1 & 87.3 & 68.7 & 57.8 & 71.4
        \\\midrule
        \multirow{7}{*}{\rotatebox{90}{\footnotesize Grouping}}& OpenPose~\cite{centernet-zhou}& \scriptsize{VGG-19}  & 61.0 & 84.9 & 67.5 & 56.3 & 69.3 & 61.8 & 84.9 & 67.5 & 57.1 & 68.2\\
        & PifPaf~\cite{pifpaf} & \scriptsize{ResNet-152}  & 67.4 & 86.9 & 73.8 & 63.1 & 74.1 & 66.7 & 87.8 & 73.6 & 62.4 & 72.9\\
        & PersonLab~\cite{personlab} & \scriptsize{ResNet-152} & 66.5 & 86.2 & 71.9 & 62.3 & 73.2 & 66.5 & 88.0 & 72.6 & 62.4 & 72.3\\
        & \multirow{2}{*}{AE~\cite{asso_embedding,higherhrnet}} & \scriptsize{HrHRNet-W32}  & 67.1	& 86.2 & 73.0 & 61.5 & 76.1 & 66.4 & 87.5 & 72.8 & 61.2&	74.2\\
        &  & \scriptsize{HrHRNet-W48}  & 69.9	& 87.2 & 76.1 & 65.4 & 76.4 & 68.4 & 88.2 & 75.1 & 64.4 & 74.2\\
        & HGG~\cite{HGG} & \scriptsize{Hourglass} & 60.4 & 83.0 & 66.2 & $-$ & $-$ & 67.6$^\dagger$ & 85.1$^\dagger$ & 73.7$^\dagger$ & 62.7$^\dagger$ & 74.6$^\dagger$ \\
        & SimplePose~\cite{SimplePose} & \scriptsize{IMHN} & 66.1 & 85.9 & 71.6 & 59.8 & 76.2 & 68.5$^\dagger$ & 86.7$^\dagger$ & 74.9$^\dagger$ & 66.4$^\dagger$ &71.9$^\dagger$\\
        \midrule
        \multirow{8}{*}{\rotatebox{90}{\footnotesize Direct}}
        & SPM~\cite{spm} & \scriptsize{Hourglass} & $-$ & $-$ & $-$ & $-$ & $-$ & 66.9 & 88.5 & 72.9 & 62.6 & 0.731\\
        & CenterNet~\cite{centernet-zhou}& \scriptsize{Hourglass}  & 64.0 & 85.6 & 70.2 & 59.4 & 72.1 & 63.0 & 86.8 & 69.6 & 58.9 & 70.4\\\cmidrule{2-13}
        & \multirow{2}{*}{\makecell[c]{DEKR~\cite{DEKR-pose}\\(w. Rescoring)}} & \scriptsize{HRNet-W32} & 68.0 & 86.7 & 74.5 & 62.1 & 77.7 & 67.3 & 87.9 & 74.1 & 61.5 & 76.1\\
        & & \scriptsize{HRNet-W48} & 71.0 & 88.3 & 77.4 & 66.7 & 78.5 & 70.0 & 89.4 & 77.3 & 65.7 & 76.9\\
        \cmidrule{2-13}
        & \multirow{2}{*}{\makecell[c]{DEKR~\cite{DEKR-pose}\\\textit{\footnotesize(w.o. Rescoring)}}} & \scriptsize{HRNet-W32} & 67.2 & 86.3 & 73.8 & 61.7 & 77.1 & 66.6 & 87.6 & 73.5  & 61.2  & 75.6 \\
        & & \scriptsize{HRNet-W48} & 70.3 & 87.9 & 76.8 & 66.3 & 78.0 & 69.3 & 89.1  & 76.7 & 65.3 & 76.4 \\\cmidrule{2-13}
        & \multirow{2}{*}{\makecell[c]{LOGO-CAP (Ours)}} & \scriptsize{HRNet-W32} & 69.6 &  87.5 & 75.9 & 64.1 & 78.0 & 68.2 & 88.7 & 74.9 & 62.8 & 76.0\\
        &                      & \scriptsize{HRNet-W48} & \textbf{72.2} &   \textbf{88.9}    &  \textbf{78.9}    &  \textbf{68.1}     &   \textbf{78.9} & \textbf{70.8} & \textbf{89.7} & \textbf{77.8} & \textbf{66.7} & \textbf{77.0}  \\
        \bottomrule
    \end{tabular}
    }
    \label{tab:coco-2017}
    \vspace{-3mm}
\end{table*}
In this section, we present detailed experimental results
and analyses of the proposed LOGO-CAP. 
\textbf{Our PyTorch source code will be released for reproducibility.}

\myparagraph{Training and Testing Settings.}
We train two LOGO-CAP networks with the ImageNet pretrained HRNet-W32 and HRNet-W48~\cite{HRNet} as the feature backbone respectively on the COCO-train-2017 dataset~\cite{COCO-dataset}. Common training and testing specifications are used in experiments, which are provided in the appendix.

\subsection{Datasets and Evaluation Metrics} \vspace{-1mm}
Two datasets are used: \textbf{The COCO dataset~\cite{COCO-dataset}} is the most popular testbed for human pose estimation. It consists of 65k, 5k and 20k images with human pose well-annotated in the training, validation and testing datasets respectively. In all experiments, the proposed LOGO-CAP is trained using the 65k training images. 
\textbf{The OCHuman dataset}~\cite{OCHuman} is one popular \textit{testing-only} dataset for evaluating human pose estimation under the occlusion scenarios. It consists of a total number of 4713 images with 8110 detailed annotated human pose instances using the COCO keypoint configuration. All the annotated 8110 human pose instances have occlusions with the maxIOU$\geq 0.5$. Furthermore, $32\%$ instances are more challenging with the maxIOU$\geq 0.75$. 

\begin{figure}[!t]
    \centering
    
    \includegraphics[height=0.18\linewidth]{figures/supp/000000006954}\hfill
    \includegraphics[height=0.18\linewidth]{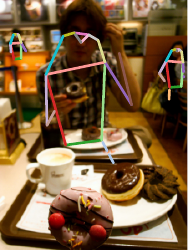}\hfill
    \includegraphics[height=0.18\linewidth]{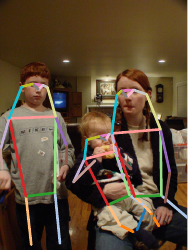}\hfill
    \includegraphics[height=0.18\linewidth]{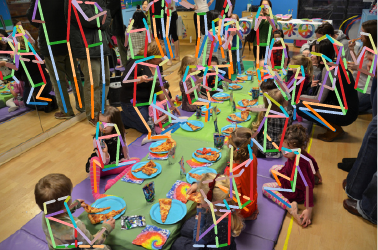}\hfill
    \includegraphics[height=0.18\linewidth]{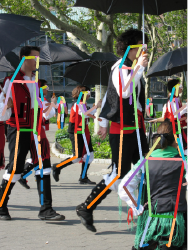}
    
    \includegraphics[height=0.18\linewidth]{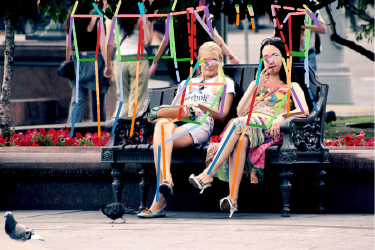}\hfill
    \includegraphics[height=0.18\linewidth]{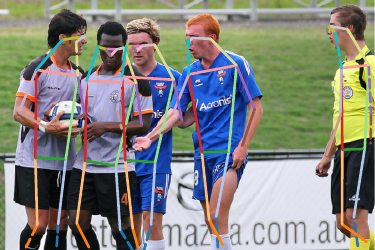}\hfill
    \includegraphics[height=0.18\linewidth]{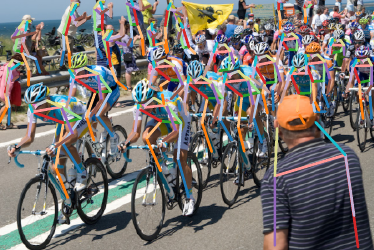}\hfill
    \includegraphics[height=0.18\linewidth]{figures/supp/000000476415}
    
    \includegraphics[height=0.18\linewidth]{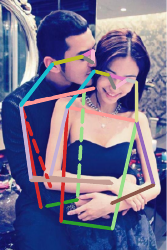}\hfill
    \includegraphics[height=0.18\linewidth]{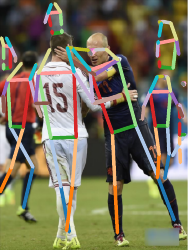}\hfill
    \includegraphics[height=0.18\linewidth]{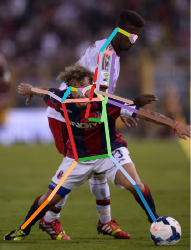}\hfill
    \includegraphics[height=0.18\linewidth]{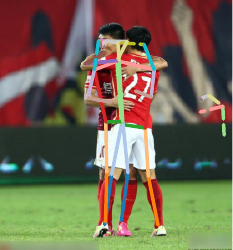}\hfill
    \includegraphics[height=0.18\linewidth]{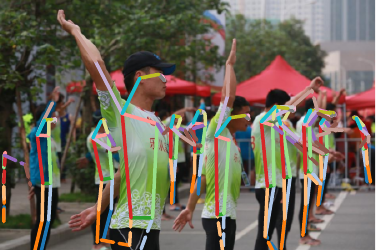}\hfill
    \includegraphics[height=0.18\linewidth]{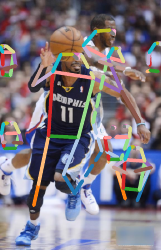}
    
    \includegraphics[height=0.18\linewidth]{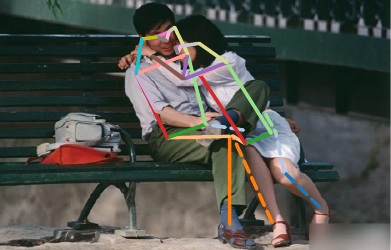}\hfill
    \includegraphics[height=0.18\linewidth]{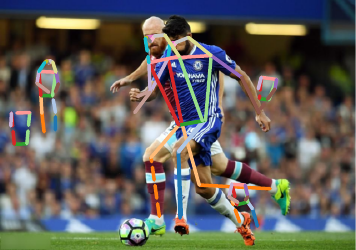}\hfill
    \includegraphics[height=0.18\linewidth]{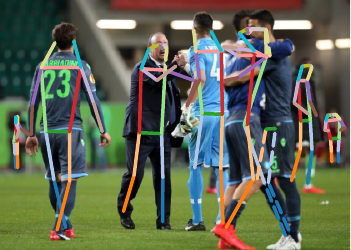}\hfill
    \includegraphics[height=0.18\linewidth]{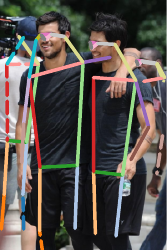}
    
    \vspace{-2mm}
    \caption{Qualitative results of our LOGO-CAP (HRNet-W32). All images were picked thematically without considering our algorithms performance. The first two rows display our approach on the COCO-val-2017 dataset and the last two ones show our results on the OCHuman test dataset.}
    \label{fig:cocoval-2017} \vspace{-4mm}
\end{figure}    

\subsection{Results on the COCO Dataset} \vspace{-2mm}

 The proposed LOGO-CAP is compared with prior arts including the bottom-up approaches of OpenPose~\cite{openpose}, PifPaf~\cite{pifpaf}, PersonLab~\cite{personlab}, AE~\cite{asso_embedding} and DEKR~\cite{DEKR-pose}, as well as the top-down approaches of Mask-RCNN~\cite{Mask_RCNN}, HRNet~\cite{HRNet}, and the light-weight LiteHRNet~\cite{Lite-HRNet}. As reported in Table~\ref{tab:coco-2017}, the proposed LOGO-CAP outperforms all the bottom-up approaches and the efficiency-toward top-down approaches (\ie, Lite-HRNet~\cite{Lite-HRNet} and Mask-RCNN~\cite{Mask_RCNN})
 on both {\tt  validation and test-dev} datasets.

Fig.~\ref{fig:cocoval-2017} shows some qualitative examples of human pose estimation by the proposed LOGO-CAP. 

In comparisons to the best-performing grouping approach, AE~\cite{asso_embedding} with a larger backbone HrHRNet-W48~\cite{higherhrnet}, our LOGO-CAP obtains competitive performance with a smaller HRNet-32 backbone, and improves the AP score with HRNet-W48 backbone on the validation and test-dev datasets by \texttt{2.3} and \texttt{2.5} points, respectively. For the fully differentiable grouping approach HGG~\cite{HGG}, our LOGO-CAP achieves better performance by a significantly large margin, more than \texttt{9.2} points on the validation set under the single-scale testing. Although the performance of HGG is improved by the multi-scale testing on the test-dev set, the performance of our LOGO-CAP is still significantly better without using the multi-scale testing scheme.

In comparisons to the direct regression based approaches, our LOGO-CAP obtains the \emph{best results} without incurring either the matching scheme used in CenterNet~\cite{centernet-zhou}  or the additional rescoring network used in DEKR~\cite{DEKR-pose}. When we disable the rescoring network for DEKR~\cite{DEKR-pose} for fair comparisons, our LOGO-CAP significantly improves the AP on the validation and testdev datasets by \texttt{2.4} points and \texttt{1.6} points respectively when HRNet-W32 is used as backbone. The larger backbone is beneficial for both DEKR and our method, which further improves the AP score of our LOGO-CAP to \texttt{72.2} and \texttt{70.8} on the validation and test-dev dataset respectively, outperforming DEKR by \texttt{1.9} and \texttt{1.5} respectively. 

In comparisons to the top-down approaches, our LOGO-CAP outperforms the end-to-end Mask-RCNN~\cite{Mask_RCNN} by 7.7 AP on the {\tt test-dev-2017} dataset. For the efficiency-toward Lite-HRNet, we improves the AP on the test-dev split from 69.7 to 70.8. Although the performance of heavy top-down HRNet~\cite{HRNet} is better than our LOGO-CAP, it should be noted that the headnet of our method is only an one-layer convolution network on the heatmap space, instead of leveraging a very deep model in the feature space of the cropped large-size images.
\subsection{Results on the OCHuman Dataset} \vspace{-2mm}
Table~\ref{tab:OCHuman} shows that our LOGO-CAP achieves the best AP performance on both the validation and testing datasets by significant margins of \texttt{2.4} and \texttt{2.2} points in comparing with the bottom-up approaches. 
For the top-down approaches, although they obtain strong AP scores on the validation split, there exists a large performance gap between the validation and testing sets. In comparisons to DEKR~\cite{DEKR-pose} (with the rescoring network), our LOGO-CAP improves the performance from 37.9 to 39.0 and from 36.5 to 38.1 on the validation and testing splits with the same backbone HRNet-W32, respectively. The similar improvement is observed when the HRNet-W48 backbone is used, outperforming both bottom-up and top-down approaches.  
\begin{table}[!t]
    \vspace{-2mm}
    \caption{Results on the OCHuman datasets~\cite{OCHuman}.}\label{tab:OCHuman}
    \vspace{-2mm}
    \centering
    \resizebox{0.8\linewidth}{!}{
    \begin{tabular}{c|c|c|c|c}
    \toprule
     & Methods & Backbone & \makecell{Val.\\AP~[\%]} & \makecell{Test\\AP~[\%]} \\\midrule
    \multirow{3}*{\rotatebox{90}{\scriptsize{Top-down}}}
    & RMPE~\cite{RMPE} & {\footnotesize Hourglass} & 38.8 & 30.7\\
    & SBL~\cite{SimpleBaseLine}  & {\footnotesize ResNet-50}& 37.8 & 30.4\\
    & SBL~\cite{SimpleBaseLine}  & {\footnotesize ResNet-152} & 41.0 & 33.3 \\
    \midrule
    \multirow{6}*{\rotatebox{90}{\scriptsize{Bottom-up}}}& AE~\cite{asso_embedding}  & {\footnotesize Hourglass} & 32.1 & 29.5\\
    & HGG~\cite{asso_embedding} & {\footnotesize Hourglass} & 35.6 & 34.8\\
    & \multirow{2}*{DEKR~\cite{DEKR-pose}} & {\footnotesize HRNet-W32} & 37.9 & 36.5\\
    &  & {\footnotesize HRNet-W48} & 38.8 & 38.2\\\cmidrule{2-5}
    & \multirow{2}*{\makecell[c]{LOGO-CAP (Ours)}}& {\footnotesize HRNet-W32} & 39.0 & 38.1\\
    & & {\footnotesize HRNet-W48} &  \textbf{41.2}  & \textbf{40.4}\\
    \bottomrule
    \end{tabular}
    } \vspace{-2mm}
\end{table}

\subsection{Inference Speed} \vspace{-2mm}
In comparing the inference speed, we test all the models on a single TITAN RTX GPU for its popularity in practice. The average inference speed, FPS (frames per second), over the 5000 images in COCO-val-2017 is used for the comparison.  For DEKR~\cite{DEKR-pose}, we re-implement their inference code with better speed obtained for fair comparisons at the algorithm level. 
For methods that have post-processing schema on CPU, only one thread is used. 
As shown in Table~\ref{tab:speed}, our LOGO-CAP runs significantly faster than PifPaf~\cite{pifpaf} and AE~\cite{asso_embedding}. The CenterNet~\cite{centernet-zhou} runs slower than DEKR and our LOCO-CAP as it requires a post-processing scheme to match the predicted offsets to the keypoints obtained from heatmaps. Comparing with DEKR, the speed improvement of our LOGO-CAP is from the lightweight design of head modules since the same backbones are used. 
For the comparisons in Table~\ref{fig:speed-accuracy}, we run the models with different resolutions of testing images. 

\begin{table}[!t]
\centering
  \vspace{-1mm}
      \caption{The single image inference speed comparison for bottom-up human pose estimation approaches.}
  \vspace{-2mm}
         \resizebox{.85\linewidth}{!}{
  \begin{tabular}{c|c|c|c|c}
  \toprule
      Method & AP [\%]& Backbone & \makecell[c]{Time $\downarrow$\\ {\footnotesize[ms]}}& FPS $\uparrow$\\\hline
      PifPaf~\cite{pifpaf} & 67.4 & {\scriptsize ResNet-152} & 213 & 4.68\\
      AE~\cite{asso_embedding,higherhrnet} &  67.1 & {\scriptsize HrHRNet-W32}& 560 & 1.78\\
      CenterNet~\cite{centernet-zhou} & 64.0 & {\scriptsize Hourglass} & 147 & 6.80\\
      DEKR~\cite{DEKR-pose} & 68.0 & {\scriptsize HRNet-W32} & 63 & 15.8\\
      DEKR~\cite{DEKR-pose} & 71.0 & {\scriptsize HRNet-W48} & 139 & 7.21\\
      \midrule
      LOGO-CAP & 69.6 & {\scriptsize HRNet-W32} & 48 & \textbf{20.7}\\
      LOGO-CAP & 72.2 & {\scriptsize HRNet-W48} & 112 & 8.95
      \\\bottomrule
      
  \end{tabular}
  }
    \label{tab:speed}
\end{table}

Furthermore, we analyze the inference time for different number of persons in the input image. As shown in Table~\ref{tab:timing}, the main bottleneck of the inference time is the backbone. For the Local-Global Contextual Adaptation module, it only takes about 10 ms on average.
\begin{table}[!t]
        \centering
        \vspace{-1mm}
        \caption{The breakdown of inference time of the proposed LOGO-CAP method. 
        For each model, we seperately report the averaged inference time across 5000 images, the averaged inference time in the images that detect only one person, the avaraged inference time in the images that have 30 persons.}
        \vspace{-2mm}
        \resizebox{0.98\linewidth}{!}{
        \begin{tabular}{c|c|ccccc}
        \toprule
                 LOGO-CAP     & $\#$Persons& Backbone &  Local KEMs & Local KAMs & Global KAMs     \\\midrule
        \multirow{3}{*}{W32} & - & \multirow{3}*{38.6 ms}    &    3.05 ms    &    2.49 ms   &   2.85 ms \\
                       & 1  &
                            & 2.39 ms & 1.14 ms & 1.12 ms\\
                       & 30 &
                            & 3.69 ms & 3.49 ms & 5.87 ms\\\midrule
        \multirow{3}*{W48} & - & \multirow{3}*{99.9 ms}    &    4.18 ms    &    3.00 ms   &   3.34 ms\\
                                      & 1 &
                                      &    3.19 ms  &    1.10 ms   & 1.07 ms\\
                                      & 30 &
                                      &  2.97 ms  & 3.59 ms & 5.97 ms \\
        
        \bottomrule
        \end{tabular}
        }
        \label{tab:timing}
    \end{table}

\subsection{Ablation Studies on COCO Validation} \vspace{-2mm}
In this section, we run a series of experiments to verify the effectiveness of the design for our proposed Local-Global Contextual Adaptation module. We use HRNet-W32 as our backbone for all ablation studies.

\begin{table}[!t]
    \centering
    \vspace{-1mm}
    \caption{Ablation studies on the three components of the LOGO-CAP: the OKS loss, the Gaussian reweighing method for heatmaps and the Attentive Normalization.}
    \vspace{-2mm}
    \resizebox{.97\linewidth}{!}{
    \begin{tabular}{c|c|c|c|c|c|c|c|c}
    \toprule
         & OKS Loss & Reweigh & AttNorm & AP   & AP$^{50}$ & AP$^{75}$ & AP$^M$  & AP$^L$ \\\hline
    baseline & -  & -  & -   & 60.0 & 84.4 & 66.4 & 54.0 & 71.1\\
    (a) &    \cmark  & -  & -   & 66.1 & 86.7 & 72.7 & 60.0 & 75.6\\
    (b) &     - & \cmark  & -   & 67.6 & 87.0 & 74.3 & 62.1 & 76.7\\
    (c) &     \cmark  & \cmark  & -   & 69.0 & 87.0 & 75.2 & 63.4 & 77.5\\
    (d) &     \cmark & - & \cmark & 65.8 & 86.8 & 72.3 & 59.3 & 75.4\\ 
    (e) &     -  & \cmark  & \cmark   & 67.5 & 86.6 & 74.1 & 62.2 & 76.7\\
    (f) &     \cmark  & \cmark  & \cmark   & 69.6 & 87.5 & 75.9 & 64.1 & 78.0\\
        \bottomrule
    \end{tabular}
    }
    \label{tab:ablation-study-1} \vspace{-2mm}
\end{table}

\myparagraph{Designs for Contextual Adaptation.}
We train 6 models to study the effectiveness in Table~\ref{tab:ablation-study-1} for the used components: (1) OKS Loss for learning local KAMs, (2) the Gaussian Reweighing scheme and (3) the Attentive Norm for convolutional message passing. Compared with the center-offset baseline, the model (a) trained with OKS loss for the local KAMs estimation obtains a large improvement of AP by 6.1 points, justifying that the potential of using local KAMs for better pose estimation. For (b), we set the factor of OKS Loss to $0$ and train the model by using the end-to-end reweighing scheme. In this setting, the local KAMs are learned only under the supervision of the contextual adaptation. Similar to model (a), a large improvement is also obtained compared against the baseline. Then, we train the model (c) that uses both OKS loss and the reweighing scheme while replacing the Attentive Norm~\cite{AttnNorm} to BatchNorm in the convolutional message passing module. It is shown that the OKS loss and the reweighing scheme collaborate very well with further improvement obtained. For the effectiveness of Attentive Norm, 
it is shown in Table~\ref{tab:ablation-study-1}~(c-f) that its feature recalibration mechanism  requires different information sources in human pose estimation. By enabling all the components, our LOGO-CAP-W32 finally obtains an AP of 69.6 on the COCO-val-2017 dataset.

\begin{table}
        \centering
        \vspace{-1mm}
        \caption{Ablation study of the different size of the local KEMs.}
        \vspace{-2mm}
        \resizebox{.97\linewidth}{!}{
        \begin{tabular}{c|c|cccccc}
        \toprule
        \makecell{size of \\ local KEM} & \makecell{Message\\ Passing} &  AP  & AP50 & AP75 &  APM & APL & FPS\\\midrule
        $7\times 7$       & 3+1 & 68.4 & 86.6 & 74.9 & 63.4 & 76.6 & 21.8\\
        $11\times 11$     & 1   & 68.8 & 86.9 &	74.9 &	63.1 &	77.5 & 22.5\\
        $11\times 11$     & 3+1   & 69.6 & 87.5 & 75.9 & 64.1 & 78.0 & 20.7\\
        $15\times 15$     & 3+1   & 69.3 & 87.1 & 75.2 & 63.2 & 78.3 & 16.5\\
        $19\times 19$     & 3+1   & 69.0 & 87.1 & 75.2 & 62.8 & 78.2 & 13.2\\
        \bottomrule
        \end{tabular}
        }
        \label{tab:ablation-study-kernel_size} \vspace{-2mm}
\end{table}

\myparagraph{Size of the Local KEMs and the Design of Convolutional Message Passing.}
We perform an ablation study with the results shown in Table~\ref{tab:ablation-study-kernel_size}, which confirms that the kernel size of $11\times 11$ obtains the best performance. One possible explanation is that smaller kernel sizes fail to compensate the uncertainty of the initial keypoint estimation results, while larger kernel sizes may introduce more nuisances factors that affect the performance, such as the ``collision" between different local KEMs of different keypoints from either the same person or adjacent different persons. Third, for different designs of the CMP module, we perform the ablation study for the different implementations. In detail, we replace the original architecture that is consist of 3 Conv+Norm+ReLU layers and 1 output Conv layer (denoted by \texttt{3+1} in Table~\ref{tab:ablation-study-kernel_size}) to a simpler architecture that only use 1 output Conv layer (denoted by \texttt{1}) for dimension reduction. It is shown that the \texttt{3+1} architecture performs better than the one only using the output Conv layer.

\begin{table}[!th]
        \centering
        \vspace{-1mm}
        \caption{Ablation study of using different priors in LOGO-CAP.} 
        \vspace{-2mm}
        \resizebox{.85\linewidth}{!}{
        \begin{tabular}{c|ccccc}
        \toprule
        Type of the Prior   &  AP   & AP50 & AP75 &  APM & APL \\\midrule
        KEMs only           &  59.4 & 80.8 & 62.8 & 50.9 & 71.6\\
        KAMs only           &  65.7 & 86.0 & 72.3 & 60.6 & 74.0\\
        \makecell{LOGO-CAP\\(KEMs + KAMs)}  & 69.6  & 87.5 & 75.9 & 64.1 & 78.0\\
        \bottomrule
        \end{tabular}
        }
        \label{tab:ablation-prior} \vspace{-2mm}
\end{table}
\myparagraph{Different Type of the Priors for Contextual Adaptation.}
As our contextual adaptation takes both the global KEMs and KAMs as priors for final pose predictions, we quantitatively compare the possible designs for the contextual adaptation. In Table~\ref{tab:ablation-prior}, we compare the performance on the COCO-val-2017 dataset by using either global KEMs or the learned global KAMs to the one that use two sources for prediction. On one hand, since the global KEMs are actually the standard Gaussian around each initial keypoint, it cannot provide more information for refinement. On another hand, when we enforce the use of local KAMs for adaptation with only global KEMs, the uncertainty from local KEMs will affect the results. That is the reason why only using global KEMs is worse than using global KAMs. When using both global KEMs and KAMs, our approach obtains the best performance.

\subsection{Potentials and Limitations} \label{sec:limitation}\vspace{-2mm}
Consider the generic applicability of the center-offset formulation to many computer vision tasks as demonstrated in~\cite{centernet-zhou}, we hypothesize that the proposed LOGO-CAP has a great potential to remedy the lack of sufficient accuracy using the vanilla center-offset method in those tasks. We also notice that the minimally-simple design in learning the contextual adaptation for refinement can be relaxed for different accuracy-speed trade-offs in practice. For example, for the convolutional message passing module, an alternative method could be the Transformer model~\cite{vaswani2017attention}, which potentially will further improve the performance at the expense of inference speed. We leave these for future work. 

\vspace{-2mm}
\section{Conclusion}\vspace{-2mm}
This paper presents a method of learning LOcal-GlObal Contextual Adaptation for Pose estimation in a bottom-up fashion, dubbed as LOGO-CAP. The proposed LOGO-CAP is built on the conceptually simple center-offset paradigm. The key idea of our LOG-CAP is to lift the center-offset predicted keypoints to keypoint expansion maps (KEMs) to address the inaccuracy of the initial keypoints. Two types of KEMs are introduced: Local KEMs are used to learn keypoint attraction maps (KAMs) via a convolutional message passing module that accounts for the structural information of human pose. Global KEMs are used to learn local-global contextual adaptation which convolves global KEMs using the KAMs as kernels. The refined global KEMs are used in computing the final human pose estimation. The proposed LOGO-CAP obtains state-of-the-art performance in COCO val-2017 and test-dev 2017 datasets for bottom-up human pose estimation. It also achieves state-of-the-art transferability performance in the OCHuman dataset with the COCO trained models.

\appendix

\section{Experimental Settings}\label{appendix:exp-setting}
Our PyTorch source code will be released. We briefly present the details of training and testing as follows. 

\myparagraph{Training.}
We train two LOGO-CAP networks with the ImageNet pretrained HRNet-W32 and HRNet-W48~\cite{HRNet} as the feature backbone respectively on the COCO-train-2017 dataset~\cite{COCO-dataset}. Common training specifications are used for simplicity in experiments. The Adam optimizer~\cite{adam-optimizer} is used with default coefficients $\beta_1=0.9$ and $\beta_2=0.999$. For both the backbones, the total number of epochs is set to $140$ and the batch size is set to $12$ images per GPU card. The same learning rate schedule is used for both models.  The learning rate is initially set to $0.001$ and then decayed to $10^{-4}$ and $10^{-5}$ at the 90-th and 120-th epoch respectively. We use $4$ and $8$ V100 GPUs to accelerate the training for the two LOGO-CAP models with HRNet-W32 and HRNet-W48 respectively. The resolution of training images is $512\times 512$ and $640\times 640$ for the two models respectively. Following the widely adopted experimental settings in~\cite{DARK}, the data augmentations in training  include (1) random rotation with the rotation degree from $-30^\circ$ to $30^\circ$, (2) random scaling with the factor in the range of $[0.75,1.5]$, (3) random translation in the range $[-40\text{pix},40\text{pix}]$ along both $x$ and $y$ directions, and (4) random horizontal flipping with the probability of $0.5$. 

Similarly, for different hyperparamters such as the trade-off parameter $\lambda$ in the total loss, we did not run computationally expensive hyperparameter optimization for simplicity. 

\myparagraph{Testing.}
We focus on the single-scale testing protocol in the COCO keypoint benchmark for the sake of efficient human pose estimation.
In the testing phase, the short side of input images is resized to a specific length (\eg, 384, 512, or 640 pixels) and keep unchanged the aspect ratio between the height and the width. As commonly adopted in many bottom-up pose estimation approaches (\eg, AE~\cite{asso_embedding}, HrHRNet~\cite{higherhrnet}, DEKR~\cite{DEKR-pose}), the flip testing is used as our default setting for the fair comparison. In the implementation, we feed the stacked tensor with an input image and a horizontally-flipped one together to get the global heatmaps and the offset fields. The flipped outputs are then averaged (according to the flip index) to get the final global heatmaps and the offset fields. For the computation of local heatmaps and the local-global adaptation, only the non-flipped outputs are used for the final predictions.

\section{The Empirical Upper Bound}

We elaborate on the details of computing the empirical upper bound of performance for a vanilla center-offset pose estimation method. The detailed results are reported in Tab.~\ref{tab:observation}. 

\begin{table}
\centering
    \caption{The performance of a vanilla center-offset regression approach, its empirical upper bound, and the performance of our proposed LOGO-CAP using HRNet-W32~\cite{HRNet} as the  feature backbone.
                  See text for detail.}\label{tab:observation}
              \begin{tabular}{c c c c}\\\toprule  
               & Baseline & Emp. Bound & LOGO-CAP \\\midrule
              AP & 60.1  & 88.9 & 70.0\\  
              AP$^{50}$ & 85.2 & 93.1 & 88.2\\  
              AP$^{75}$ & 66.7 & 90.6 & 76.4\\  
              AP$^{M}$ &  53.7 & 87.7 & 64.4\\ 
              AP$^{L}$ &  71.5 & 90.2 & 78.4\\
              \bottomrule 
              \end{tabular}
\vspace{-2mm}
\end{table}
\myparagraph{Network Architecture.} The vanilla center-offset regression baseline uses the ImageNet pretrained HRNet-W32~\cite{HRNet} as the backbone, and the same modules as in our LOGO-CAP+HRNet-W32 for the center heatmap regression and the offset vector regression. We present the details of computing keypoint expansion maps (KEMs) that are used in calculating the empirical uppper bound as follows.   

\myparagraph{Computation of Keypoint Expansion Maps.} Denoted by $\mathcal{P} \in \mathbb{R}^{N\times 17 \times 2}$ the initial pose parameters (i.e., the 2-D locations for the $17$ keypoints of the $N$ pose instances) estimated by the vanilla center-offset method, we expand each of the estimated keypoints with a local $11\times 11$ mesh grid, that is to lift a keypoint to a 2-D mesh to counter the estimation uncertainty. We use the COCO benchmark provided keypoint sigmas to scale the unit length of the meshgrid for different types (\eg, nose, eyes, hips) of keypoints. After getting the expanded keypoint meshes $\mathcal{M}$ of the initial poses, we compute their keypoint similarities $\mathcal{S}\in\mathbb{R}^{N\times 11\times 11\times K \times 17}$ between the groundtruth keypoints $\mathcal{G}\in\mathbb{R}^{K\times 17\times 3}$ and the keypoint expansion maps. By applying the sum reduction on the similarity tensor $\mathcal{S}$ along the $2$-nd, $3$-rd and the last axes, we have known the optimal correspondence  (including the low-quality matches) for each center anchor, denoted by $\mathcal{S}_{N\times 11\times 11\times 17}$. Then, the pose with the maximal similarity in the $11\times 11$ local window for each center anchor are used as the best one to compute the empirical upper bound on the fully-annotated COCO-val-2017 dataset.

\section{More Qualitative Results}
\begin{figure}[!h]
    \centering
    
    \includegraphics[height=0.15\linewidth]{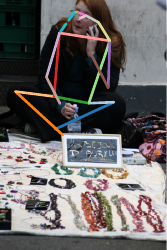}\hfill
    \includegraphics[height=0.15\linewidth]{figures/supp-cvpr/coco/000000508730.pdf}\hfill
    \includegraphics[height=0.15\linewidth]{figures/supp-cvpr/coco/000000545594.pdf}\hfill         \includegraphics[height=0.15\linewidth]{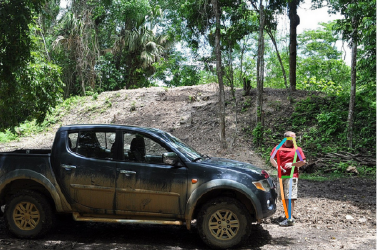}\hfill
    \includegraphics[height=0.15\linewidth]{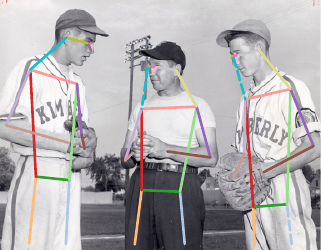}\hfill
    
    \includegraphics[height=0.15\linewidth]{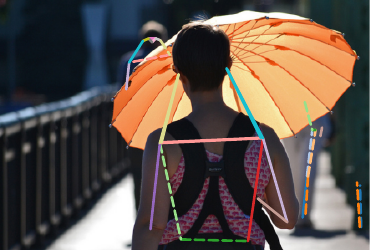}\hfill
    \includegraphics[height=0.15\linewidth]{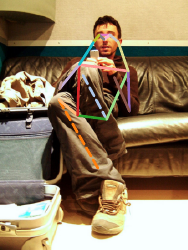}\hfill
    \includegraphics[height=0.15\linewidth]{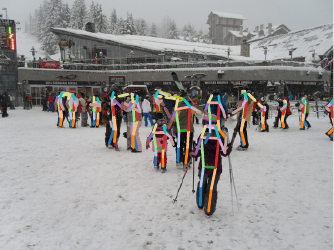}\hfill
    \includegraphics[height=0.15\linewidth]{figures/supp-cvpr/coco/000000192607.pdf}\hfill
    \includegraphics[height=0.15\linewidth]{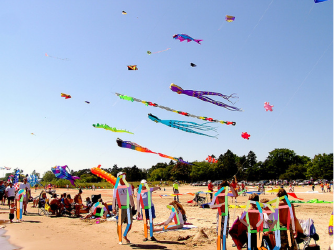}
    
    \includegraphics[height=0.16\linewidth]{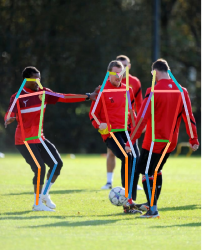}\hfill
    \includegraphics[height=0.16\linewidth]{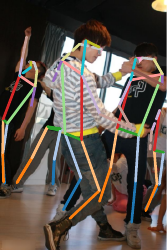}\hfill
    \includegraphics[height=0.16\linewidth]{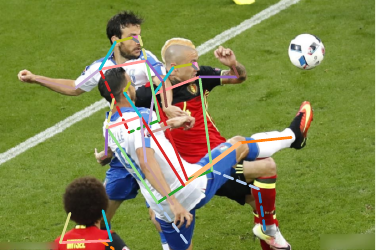}\hfill
    \includegraphics[height=0.16\linewidth]{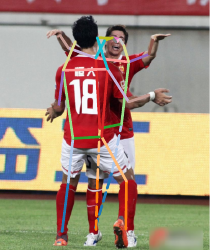}\hfill
    \includegraphics[height=0.16\linewidth]{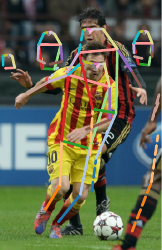}\hfill
    \includegraphics[height=0.16\linewidth]{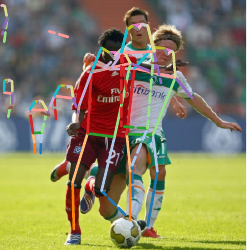}\hfill
    
    \includegraphics[height=0.16\linewidth]{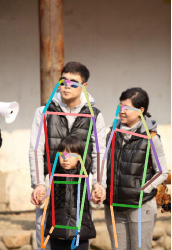}\hfill
    \includegraphics[height=0.16\linewidth]{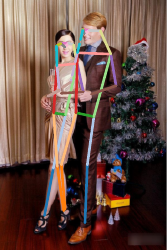}\hfill
    \includegraphics[height=0.16\linewidth]{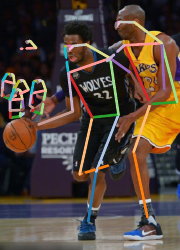}\hfill
    \includegraphics[height=0.16\linewidth]{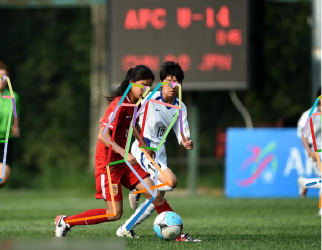}\hfill
    \includegraphics[height=0.16\linewidth]{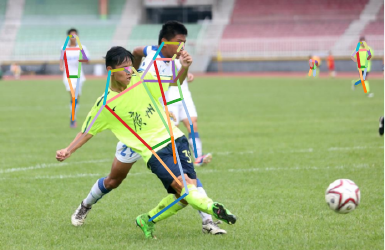}
    
    \caption{Qualitative results of our LOGO-CAP (HRNet-W32). All images were picked thematically without considering our algorithms performance. The first two rows display our approach on the COCO-val-2017 dataset and the last two ones show our results on the OCHuman test dataset.}
    \label{fig:cocoval-2017-supp}
\end{figure}

\myparagraph{Results on the COCO-val-2017 and the OCHuman Datasets.} Fig.~\ref{fig:cocoval-2017-supp} shows examples of pose estimation in the two datasets by the proposed LOGO-CAP with the HRNet-W32 backbone. Our proposed LOCO-CAP is able to handle large structural and appearance variations in human pose estimation. 

\myparagraph{Fast pose estimation for video frames.} To justify the potential of our proposed approach in practical applications, we run our LOGO-CAP (W32 model) on two videos that have the resolution of $1280\times 720$ from YouTube. We follow our testing protocol to resize the short side of the video frames to $512$ pixels and keep their original aspect ratios for inference. 
Without using any pose tracking techniques, our LOGO-CAP achieves fast and accurate human pose estimation. Please click the following anonymous links for the demo videos (\textbf{with background musics}):
\begin{itemize}
    \item[-] \url{https://bit.ly/3cFcJ75} (video credit: \url{https://youtu.be/2DiQUX11YaY})
    \item[-] \url{https://bit.ly/30RkyEg} (video credit: \url{https://youtu.be/kTvzU1sGSyA})
\end{itemize}
In these two demo videos, the instantaneous FPS for each video frame is marked in the left corner of the video.

{\small
\bibliographystyle{ieee_fullname}
\bibliography{egbib}
}

\end{document}